\documentclass[letterpaper, 10 pt, conference]{ieeeconf} 

\usepackage{cite}

\usepackage{amsmath,amsfonts}
\usepackage{algorithmic}
\usepackage{algorithm}
\usepackage{array}
\usepackage[caption=false,font=normalsize,labelfont=sf,textfont=sf]{subfig}
\usepackage[colorlinks=true,linkcolor=blue, urlcolor=black, citecolor=cyan]{hyperref}
\usepackage{textcomp}
\usepackage{stfloats}
\usepackage{url}
\usepackage{verbatim}
\usepackage{graphicx}
\usepackage{cite}
\usepackage[acronym]{glossaries}

\usepackage[table,xcdraw]{xcolor}
\usepackage[caption=false]{subfig}
\usepackage{multirow}\usepackage{graphicx}
\begin{document}
%
\title{MorphoGear: An UAV with Multi-Limb Morphogenetic Gear for Rough-Terrain Locomotion}


\author{Mikhail Martynov, Zhanibek Darush, Aleksey Fedoseev, Dzmitry Tsetserukou

\thanks{Mikhail Martynov, Zhanibek Darush, Aleksey Fedoseev, Dzmitry Tsetserukou are with the ISR Laboratory, Skolkovo Institute of Science and Technology (Skoltech), 143026 Moscow, Russia.
(e-mail: mikhail.martynov@skoltech.ru; zhanibek.darush@skoltech.ru;
aleksey.fedoseev@skoltech.ru; d.tsetserukou@skoltech.ru).}

\thanks{}
\thanks{}}

%
%

\markboth{TMECH/AIM Focused Section Submission}%
{}
%



\maketitle

\begin{abstract}
Robots able to run, fly, and grasp have a high potential to solve a wide scope of tasks and navigate in complex environments. Several mechatronic designs of such robots with adaptive morphologies are emerging. However, the task of landing on an uneven surface, traversing rough terrain, and manipulating objects still presents high challenges.

This paper introduces the design of a novel rotor UAV MorphoGear with morphogenetic gear and includes a description of the robot's mechanics, electronics, and control architecture, as well as walking behavior and an analysis of experimental results. MorphoGear is able to fly, walk on surfaces with several gaits, and grasp objects with four compatible robotic limbs. Robotic limbs with three degrees of freedom (DoFs) are used by this UAV as pedipulators when walking or flying and as manipulators when performing actions in the environment. 
We performed a locomotion analysis of the landing gear of the robot. Three types of robot gaits have been developed.

The experimental results revealed low crosstrack error of the most accurate gait (mean of 1.9 cm and max of 5.5 cm) and the ability of the drone to move with a 210 mm step length. Another type of robot gait also showed low crosstrack error (mean of 2.3 cm and max of 6.9 cm). The proposed MorphoGear system can potentially achieve a high scope of tasks in environmental surveying, delivery, and high-altitude operations.

\end{abstract}


%
\IEEEpeerreviewmaketitle

\section{Introduction}
%
%
%
%

\IEEEPARstart{I}{n} recent years, the impact of unmanned aerial vehicles (UAVs) has continued to increase in a wide scope of moni-toring and inspection tasks. High mobility allows drones to overcome dense obstacles and collect data from areas inaccessible to unmanned ground vehicles (UGVs). Intelligent UAVs possess much higher capabilities for autonomous inspections, adapting to dynamic, uncertain environments and complex tasks, and collecting data from extensive areas. However, the low capacity of the power supply and the low payload do not allow for extensive use of rotor UAVs for long-term inspections. Moreover, indoor inspections often require the ability of the robot to overcome dense obstacles and operate in environments on the ground and in midair.
\begin{figure}[!h]
\centering
\includegraphics[width=1\linewidth]{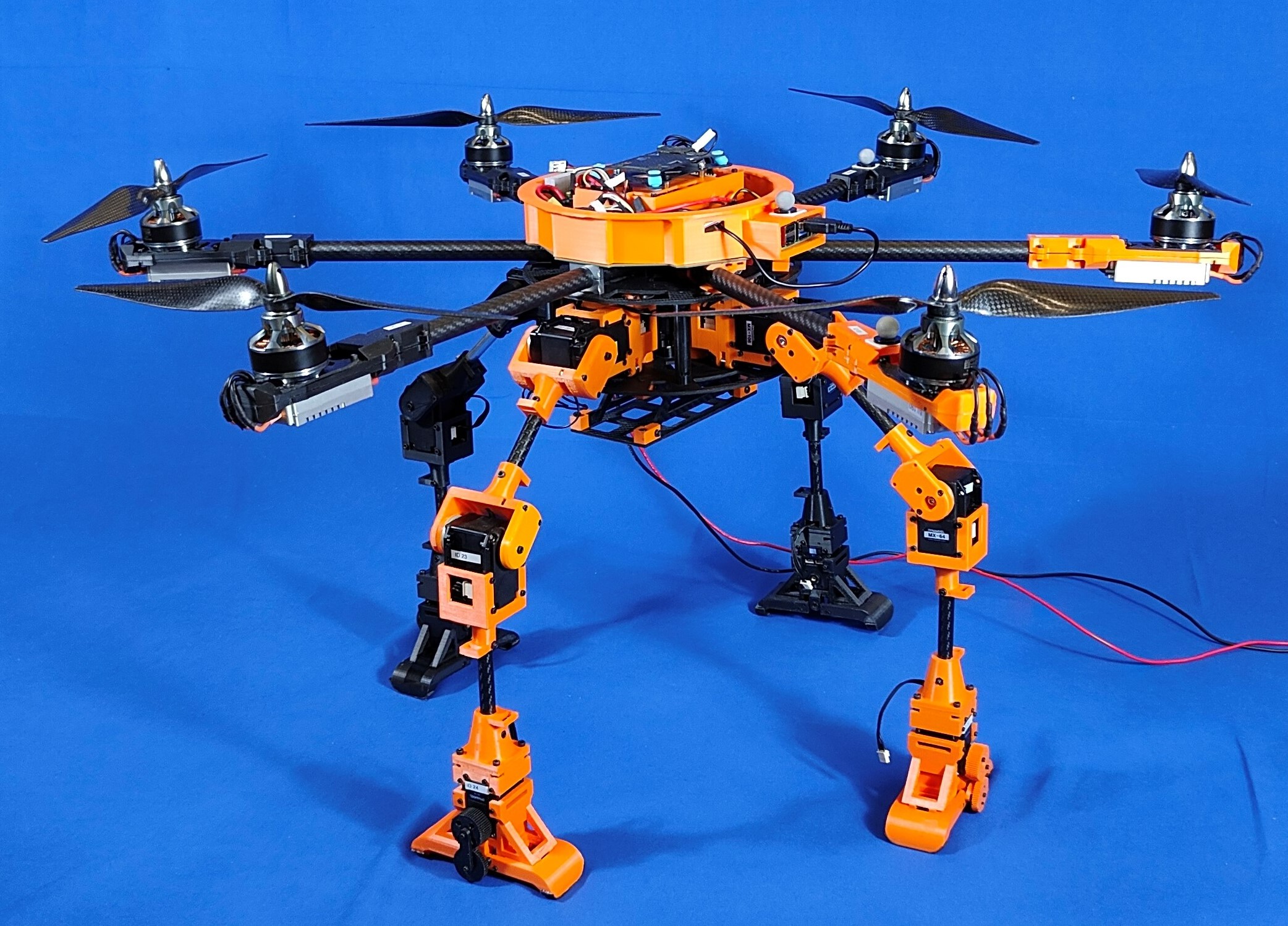}
\caption{Multi-limb morphogenetic UAV for aerial manipulation and terrestrial locomotion}
\label{fig_main}
\end{figure}
To overcome the shortcomings of UAVs, several researchers suggested integra-ting UAV and UGV designs in flying robots with adaptable morphology, enabling them to alternate between different modes of movement~\cite{Zhang_2022}.

Recent studies in the field of morphogenetic robotics explore the ability of adaptive morphogenesis to increase the efficiency of using robots in dynamic environments. Daler et al. \cite{Daler_2015} introduced the concept of morphogenetic fixed-wing UAVs with wings able to transform between flying and crawling modes. Later, Mulgaonkar \cite{Mulgaonkar_2016} developed a design based on a nanocopter that could fly, crawl with six limbs actuated synchronously by a single DC motor, and grasp objects of fixed size with these limbs. However, the grasping ability of this UAV is highly limited. Sun et al. \cite{Sun_2019} proposed another mechatronic concept for walking robots with shape morphing joints. A morphing robotic platform traversing land and water with six spherical footpads was proposed by Kim et al. \cite{Kim_2016}. Bio-inspired design for the similar problem of motion on water and ground was suggested by Baines et al. \cite{Baines_2022}, achieving locomotion with four morphogenetic legs similar to the turtle limbs. Another bio-inspired design was suggested for the aerial-aquatic hybrid robot developed by Li et al. \cite{Li_2022}, where a multirotor UAV was enhanced by a robotic lamellar mechanism.

This work proposes a new concept of autonomous monitoring and inspection, with the morphogenetic UAV (Fig.~\ref{fig_main}) moving through terrain with the help of four pedipulators and using them for manipulation.
In this paper, we present the mechanics and architecture of the platform. And we are also considering several options for the robot's gait.

\section{Related Works}
The scope of the morphogenetic UAVs applications relies on the ability of the drone to switch between ground locomotion and aerial exploration by landing on four legs equipped with robotic grippers. Our work is inspired by several previous papers exploring UAV's ability to interact with the environment by robotic arms and chassis. Yashin et al.~\cite{Locogear} proposed a LocoGear, a novel algorithm for locomotion of UAV equipped with the robotic landing gear.


Sarkisov et al.~\cite{SAM} presented a novel system cable-Suspended Aerial Manipulator (SAM). The SAM is equipped with two actuation systems: winches and propulsion units.
 Appius et al.~\cite{Appius_2022} proposed a robotic gripper for rapid aerial pickup and transport. Bio-inspired perching mechanism with soft grippers was proposed for flapping-wing UAVs by Broers et al.~\cite{Broers_2022}. Brunner et al.\cite{Brunner_2022} introduced a planning-and-control framework for aerial manipulation of articulated objects. Aerial manipulation through VR interface and AeroVR drone equipped with 4-DoF robotic arm was proposed by Yashin et al.~\cite{Yashin_2019}.

Apart from grasping and manipulating objects, robotic pedipulators can work as a locomotion system when integrated with UAV. Several lightweight legged platforms were previously introduced, e.g., Camacho-Arreguin et al. \cite{Camacho-Arreguin_2022} proposed a walking tool to achieve a higher efficiency, a better load capacity. A hexapod robot developed by Cizek et al. \cite{Cizek_2018} and ALPHRED four-legged platform developed by Hooks et al. \cite{Hooks_2018}, that can be potentially implemented for hybrid UAV locomotion.

Several prior projects focused on hybrid UAV systems that could also traverse terrain. For example, Kalantari et al. \cite{Kalantari_2013} presented a micro-UAV HyTAQ equipped with a rolling cage. Latscha et al. \cite{Latscha_2014} proposed to integrate two snake robots with a quadrotor using a magnetic docking system in the H.E.R.A.L.D. hybrid robot. Wheeled gear integrated with four 1-DoF limbs was proposed for ground locomotion by Ceccarelli et al. \cite{Ceccarelli_2018} for service robots in cultural heritage frames.
Pitonyak \cite{Pitonyak_2017} developed a Hexwalker robot with hexacopter and hexapod platforms. However, the Hexwalker robot does not have additional sensors. The concept is further improved by DroneGear hybrid drone developed by Sarkisov et al. \cite{Sarkisov_2018}. Kim et al \cite{Kim_2021} developed a bipedal robot design, Leonardo, able to walk, fly, and maintain balance by synchronized activation of two robotic limbs and UAV rotors.

The MorphoGear system, inspired by the previous concept of the DroneGear, proposes a new walking, flying, and grasping platform with 12 DoFs and higher accuracy of locomotion. In contrast with prior four-legged systems, the MorphoGear can utilize two-finger grippers on each limb for object manipulation.  

\section{System Overview}
\subsection{Mechanics}
\subsubsection{UAV design concept}
The robot has four limbs mounted on a single base for terrestrial locomotion and six rotors directed at 90 degrees to the ground  for flight. The distance between the opposite rotors is 800 mm, which allows to install 13-inch blades on each rotor. The weight of the entire robot is 10.4 kg. Base parts and limb tubes are made of carbon, which are connected to each other by 3D printed parts made of PLA plastic. 
The robot's symmetrical design allows it to move in any direction, without affecting the algorithms, according to the operator's choice.

\subsubsection{Drone chassis design}
\begin{figure}[htb!]
\centering
 \includegraphics[width=1\linewidth]{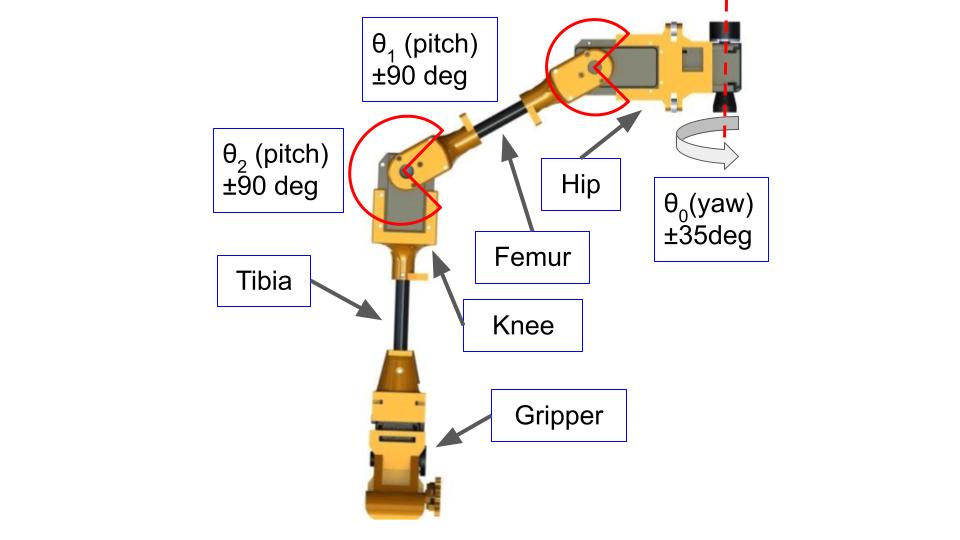}
 \caption{CAD design of the 3-DoF drone’s pedipulator with robotic gripper actuated by four Dynamixel servomotors.}
 \label{fig:leg}
 \vspace{-0.2em}
\end{figure}

The landing platform consists of four limbs with 3-DoFs, which are located in increments of $90^{\circ}$ relative to the central axis of the robot. The motion of the legs is executed by Dynamixel MX-106 and MX-28 servomotors in the hip joints and Dynamixel MX-64 servomotors in the knee joints.
The CAD model of the limb is shown in Fig.~\ref{fig:leg}. 

\subsection{Electronics and Software}
\subsubsection{General architecture}

\begin{figure}[htb!]
\centering
 \includegraphics[width=1\linewidth]{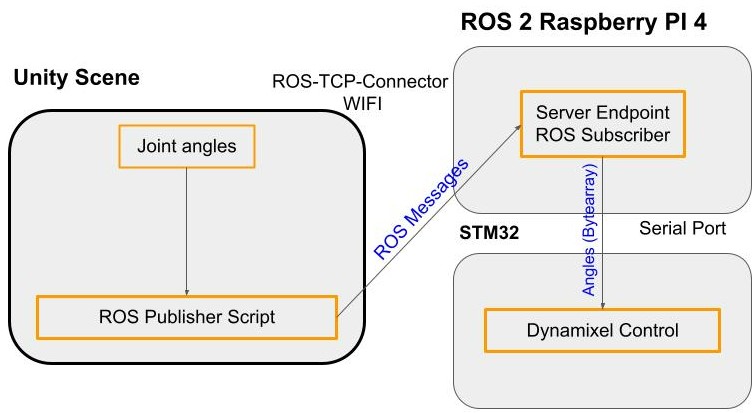}
 \caption{System software architecture. The pipeline consists of Unity Engine environment on the control PC for the calculation of the robot kinematics and ROS 2 framework providing remote connection to the robot and motor control.}
 \label{fig:architecture}
 \vspace{-0.2em}
\end{figure}

The MorphoGear software system consists of three modules: the Unity Engine simulation environment, the STM32 microcontroller as the middleware for the motor control, and the Raspberry Pi 4 as the command receiver through the Robot Operating System (ROS). Unity provides simulation of robot motion control with visualization. The control pipeline for the MorphoGear is shown in Fig.~\ref{fig:architecture}.

Commands from Unity are received by ROS using the ROS TCP Connector provided by Unity Technologies, then ROS sends commands to STM32. STM32 can read data from the encoders and control Dynamixel servomotors. STM32 returns feedback to Unity with actual information about current parameters of servomotors (voltage, rotation angle, and temperature). All calculations of the robot's movement are implemented on the Unity Game Engine using asynchronous methods. Then data of the angular position of servomotors are sent as commands.

\begin{figure}[!h]
\centering
\includegraphics[width=1\linewidth]{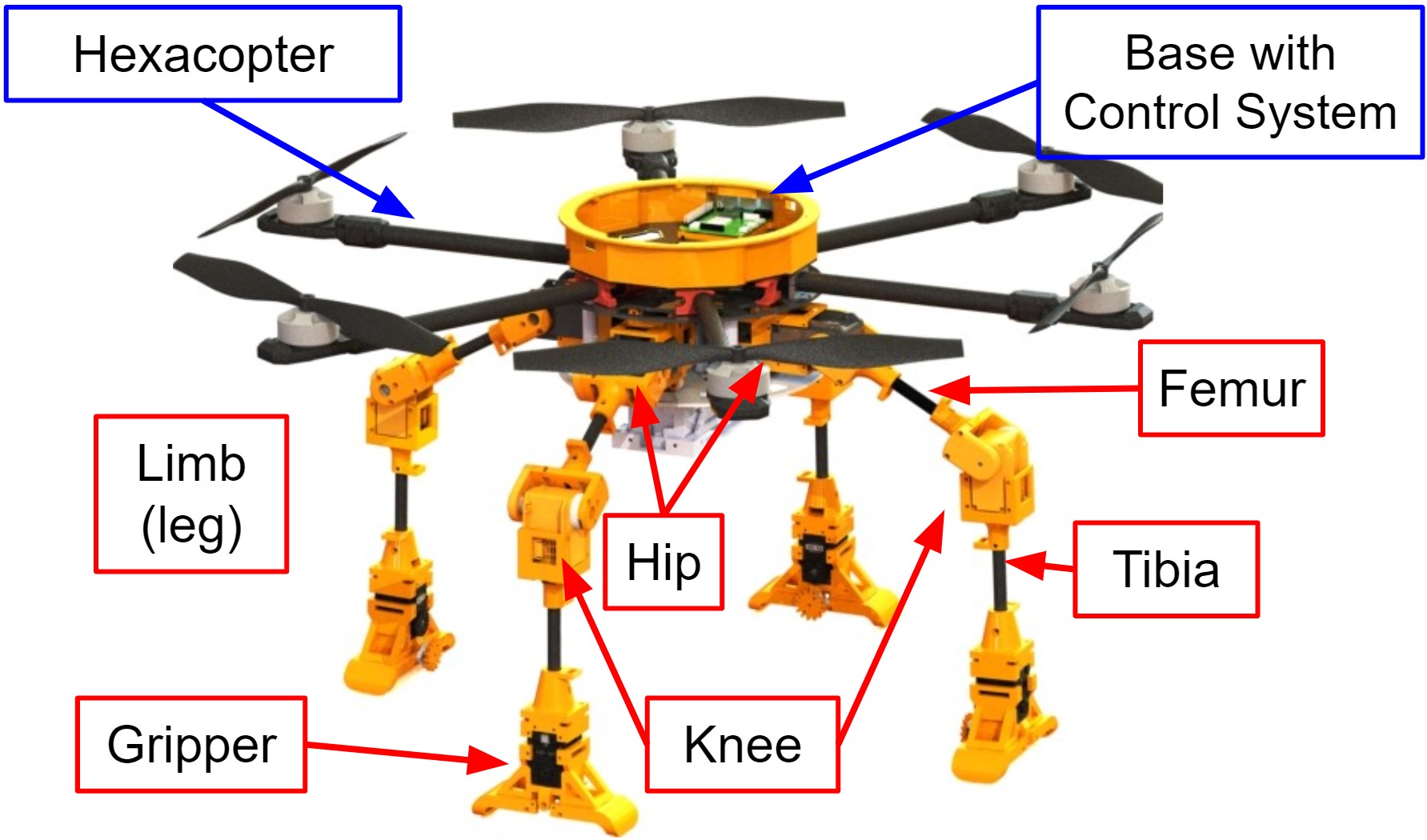}
\caption{MorphoGear CAD Rendered Model illustrating its workspace during manipulation.}
\label{fig_workspace}
\end{figure}

\subsubsection{Power supply}
The robot uses a LiPo battery located in a special compartment below the chassis. The total battery capacity is 8000 mAh, which is enough to work in flight mode for up to 12 minutes.

\section{MorphoGear Locomotion}
\subsection{Trot Gait}

The robot's walking algorithm is based on lifting itself up with two supporting limbs that are perpendicular to the ground at the last link. The MorphoGear step has two parts: moving two limbs forward and placing them on the ground, and then lifting the robot on these limbs and moving forward by rotating the hip joints (yaw).

To determine how high the robot should lift itself during each step, the horizontal shift caused by a single hip joint yaw rotation is calculated. The limb is assumed to be fixed at the point of contact with the ground and can only rotate around the axis perpendicular to the surface.

When the robot turns around the fixed point of contact, it describes part of a circle, which compresses the base of the robot during the step because the limbs are opposite each other. To avoid this, the robot lifts itself to compensate for the horizontal shift. The equation of motion is written as follows:



\begin{equation}
 \label{shift_horizontal}
 \Delta = 2(1-\cos\theta_0)l
\end{equation}
where $\theta_0$ is the hip joint angle (yaw) of the opposite supporting limbs, increasing during the first half of the step and then decreasing, $l$ is the distance from the attachment of the limb to the base to the point of touching the ground along the horizontal axis.
Thus, the robot lifts itself up for half of the movement, then lowers itself to its original position for the second half of the movement.

Lifting the robot compensates for the shift as follows: 
\begin{equation}
 \label{shift_vertical}
 \Delta = (\cos \theta_1^{init} - \cos (\theta_1^{init}+\xi)) \cdot l_{F}
\end{equation}
where $\theta_1^{init}$ is the initial angle of the hip joints(pitch), $\xi$ is the shift compensation angle, $l_{F}$ is the length of femur.
Assuming that the robot rests on its limbs perpendicular to the ground, relative $l$ is calculated as $l_{F}\cos \theta_1^{init}$. 

From equations (\ref{shift_horizontal}) and (\ref{shift_vertical}) we obtained the relationship between the angle of rotation of the leg (hip joint yaw) and the angle at which to lift the robot (hip joint pitch):
\begin{equation}
 \label{comparizon_equation}
 2(1-\cos\theta_0)l = (\cos \theta_1^{init} - \cos (\theta_1^{init} + \xi)) \cdot l_{F}
\end{equation}

Thus, by expressing the lifting angle from the equation (\ref{comparizon_equation}), we were able to obtain forward motion from the movements of two opposite limbs around circles.
\begin{equation}
 \label{xi}
 \xi = \arccos((2\cos \theta_0 - 1) \cdot \cos \theta_1^{init}) -\theta_1^{init}
\end{equation}

The limbs located in front and behind are raised at the moment of the step, so as not to interfere with the movement of the drone.

\subsection{Canter Gait}

\begin{figure}[!h]
\centering
\includegraphics[width=0.9\linewidth]{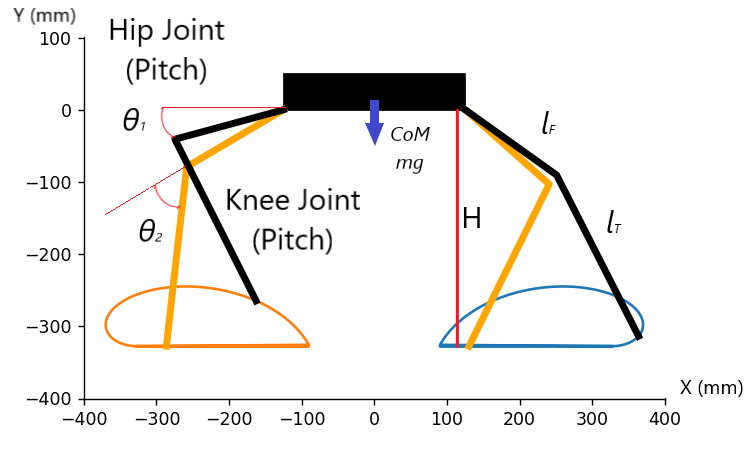}
\caption{Limb trajectories during a single step in the platform coordinate system.}
\label{fig_crab}
\end{figure}

The idea behind the following gait is that all four limbs of the robot move synchronously along an annular trajectory, the lower part of which passes at ground level and is straight, and the upper part of which is part of the Archimedes spiral. Since the mechanics of the robot are symmetrical, the trajectories for each limb are the same, but they are oppositely directed for the front and rear limbs. In addition, while the front left limb is at the beginning of the trajectory, the front right leg is at the same moment in the middle of the trajectory. Similarly, for the hind limbs.
To design the gait, knowing the height of the robot, we solve the inverse kinematics problem to find a straight line segment at ground level serving as the lower part of the trajectory, and then close the segment with an Archimedes spiral serving as the upper part of the trajectory.
The beginning of the trajectory can be considered the beginning of the segment, and the beginning of the spiral can be considered the middle of the trajectory. Thus, if the left forward limb is at the beginning of the trajectory, then the right forward limb is in the middle of the trajectory. And vice versa for the hind limbs. Thus, the front right and hind left limbs move along a segment of the same length along the ground, while the other two limbs move along a spiral forward. (Fig.~\ref{fig_crab})

Robot height can be obtained by the following equation:
\begin{equation}
 \label{robot height}
 H = \sin\theta_1^{init} \cdot l_{F} + l_{T}
\end{equation}
where $l_{T}$ is the length of the tibia.
Further, to find the beginning of the segment, it is assumed that the extreme position is achievable with a fully outstretched limb.
The end of the segment is considered to be the projection of the attachment point of the limb to the base to prevent the robot from turning over.
The beginning of the segment can be obtained by the following equation:
\begin{equation}
 \label{segment beginning}
 x_0 = \sqrt{(l_{F}+l_{T})^2-H^2}
\end{equation}
where $x_0$ is the coordinate of the beginning of the segment along the axis of movement of the limb, and the projection of the attachment point of the limb to the base considered as zero point.

Inverse kinematics of the robot limbs was calculated as follows: 
\begin{equation}
 \label{Inverse Kinematics}
 \cos\theta_2 = (x^2+y^2- l_{F}^2-l_{T}^2)/(2l_{F}l_{T}),
\end{equation}
where $x, y$ are the coordinates of the end effector relative to the base frame of the limb. Then $\theta_2$ can be extracted from this equation as follows:
\begin{equation}
\theta_2 = -\arccos(\cos(\theta_2)),
\end{equation}
\begin{equation}
 \sin\theta_2 = \sqrt{(1-\cos\theta_2^2)};
\end{equation}

Then $\beta$ is obtained according to the following equation:
\begin{equation}
\theta_1 = arctan2(y,x) + arctan2(l_{T}\sin\theta_2,l_{F}+l_{T}\cos\theta_2)
\end{equation}


\subsection{Gallop Gait}
This type of gait is a combination of the two previous gaits. The idea of this gait is very similar to the first type of walking; however, in the process of moving the two supporting limbs forward, the front and rear limbs continue to move the robot as in the second type of walking. When the robot lifts itself on two supporting limbs, as in the first type of walking, the front and rear limbs pass along the upper part of the trajectory of the second type. The advantage of a mixed gait over the first type of gait is that the robot moves continuously, and the walking speed increases. In addition, this gait allows the robot to walk strictly in a straight line and reduces the rocking of the platform, unlike the second type of walking.




\section{Experimental Evaluation}
A series of experiments were conducted to assess the quality of MorphoGear walking gaits. The Vicon MOCAP system was used to record the path of the robot along its linear and angular trajectories. Three different robot gaits were tested, with an initial $\theta_1^{init}$ value of $30^{\circ}$. The control signal was fed to the platform by transmitting angles for servomotors at a frequency of 0.005 Hz. Each gait was recorded in seven trials, with the initial standard deviation of a position not higher than 0.48 cm.
The experiment consisted of a series of trials where the robot had to walk 1 meter forward following a linear trajectory. Based on the results of the experiments, the deviation from the trajectory was analyzed.

\subsubsection{Trot Gait Evaluation}
\begin{figure}[!h]
\centering
\includegraphics[width=0.9\linewidth]{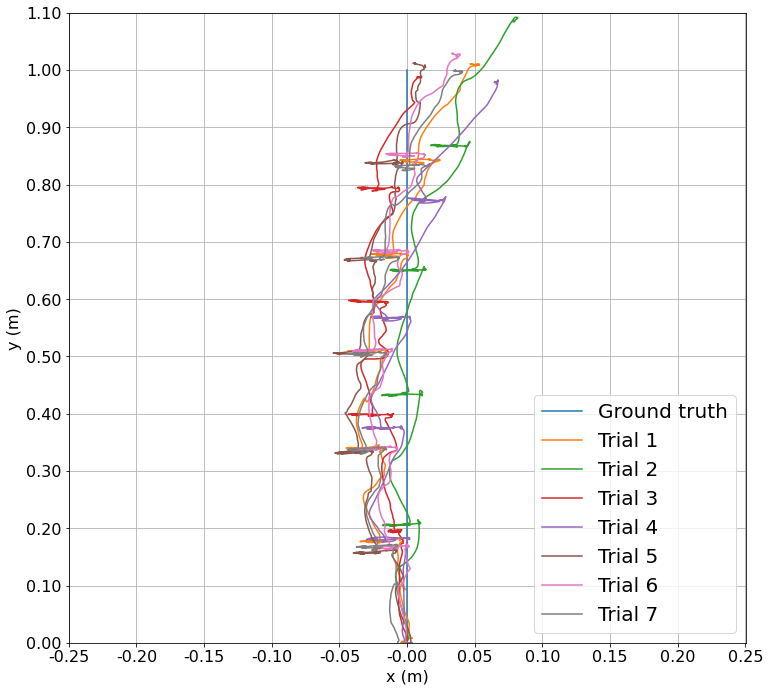}
\caption{Experimental trajectories obtained for trot gait and linear trajectory following. The light blue line indicates ground truth trajectory.}
\label{fig_type1}
\end{figure}

The trajectory of the robot's motion with the trot gait is shown in Fig.~\ref{fig_type1}.

 Table \ref{Table Result 1} shows that the RMSE does not exceed 3 cm and maximal error does not exceed 8.5 cm. 

\begin{table}[h!]
\centering{
\caption{Path Following Error \\ for Straight Line Trajectory and Trot Gait}
\setlength{\tabcolsep}{6pt} 
\renewcommand{\arraystretch}{1}
\begin{tabular}{ | c | c | c | }
\hline
\label{Table Result 1}
\begin{tabular}{p{1.5cm}cp{4cm}} \end{tabular} & {RMSE, cm} & {Max Error, cm} \\\hline
Trial 1 & 1.9 & 6.7 \\\hline
Trial 2 & 1.9 & 8.2 \\\hline
Trial 3 & 1.4 & 4.3 \\\hline
Trial 4 & 2.6 & 5.5 \\\hline
Trial 5 & 1.8 & 4 \\\hline
Trial 6 & 2.1 & 4.4 \\\hline
Trial 7 & 1.8 & 5.4 \\\hline
Overall & 1.9 & 8.2 \\\hline
\end{tabular}}
\end{table}

Additionally, the results show a low deviation between mean error in all trials, however, there is a consistent inclination of the robot to the right direction, visible in all trials.

\subsubsection{Canter Gait Evaluation}
\begin{figure}[h!]
\centering
\includegraphics[width=0.9\linewidth]{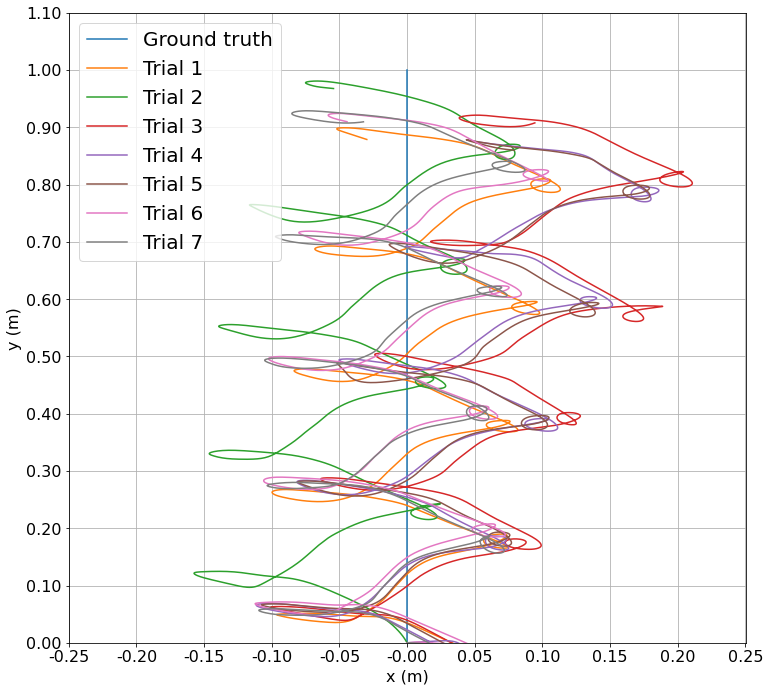}
\caption{Experimental trajectories obtained for canter gait and linear trajectory following. The light blue line indicates ground truth trajectory.}
\label{fig_type2}
\end{figure}

The trajectory of the robot's motion with the canter gait is shown in Fig.~\ref{fig_type2}.

Table \ref{Table Result 2} shows the RMSE of 6.2 cm and maximal error of 21.1 cm after one meter of linear path following. 

\begin{table}[h!]
\centering{
\caption{Path Following Error \\ for Straight Line Trajectory and Canter Gait}
\setlength{\tabcolsep}{6pt} 
\renewcommand{\arraystretch}{1}
\begin{tabular}{ | c | c | c | }
\hline
\label{Table Result 2}
\begin{tabular}{p{1.5cm}cp{4cm}} \end{tabular} & {RMSE, cm} & {Max Error, cm} \\\hline
Trial 1 & 5.1 & 11.3 \\\hline
Trial 2 & 5.9 & 15.7 \\\hline
Trial 3 & 8.2 & 21.1 \\\hline
Trial 4 & 7.2 & 18.6 \\\hline
Trial 5 & 6.9 & 17.9 \\\hline
Trial 6 & 5.1 & 11.2 \\\hline
Trial 7 & 4.9 & 10.9 \\\hline
Overall & 6.2 & 21.1 \\\hline
\end{tabular}}
\end{table}
The experimental results showed a lower path following accuracy in all trials compared with trot gait due to the UAV's center of mass (CoM) swinging when transferring its weight between the legs in each step.

\subsubsection{Gallop Gait Evaluation}
\begin{figure}[h!]
\centering
\includegraphics[width=0.9\linewidth]{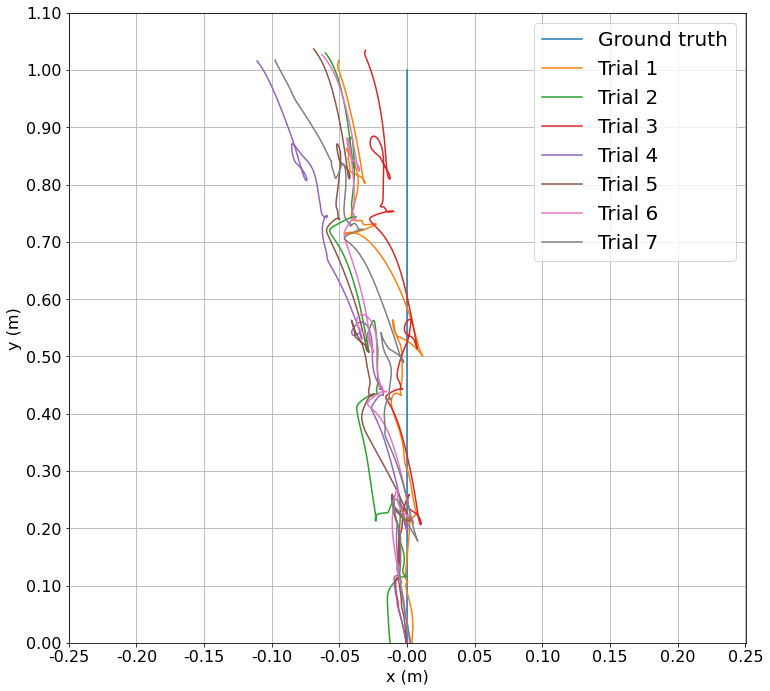}
\caption{Experimental trajectories obtained for gallop gait and linear trajectory following. The light blue line indicates ground truth trajectory.}
\label{fig_type3}
\end{figure}

The trajectory of the robot's motion with the gallop gait is shown in Fig.~\ref{fig_type3}.

Table \ref{Table Result 3} shows the RMSE of 2.3 cm and maximal error does of 11.1 cm after one meter of linear path following. 

\begin{table}[h!]
\centering{
\caption{Path Following Error \\ for Straight Line Trajectory and Gallop Gait}
\setlength{\tabcolsep}{6pt} 
\renewcommand{\arraystretch}{1}
\begin{tabular}{ | c | c | c | }
\hline
\label{Table Result 3}
\begin{tabular}{p{1.5cm}cp{4cm}} \end{tabular} & {RMSE, cm} & {Max Error, cm} \\\hline
Trial 1 & 1.6 & 5.1 \\\hline
Trial 2 & 2.7 & 6 \\\hline
Trial 3 & 1 & 3.1 \\\hline
Trial 4 & 3.5 & 11.1 \\\hline
Trial 5 & 2.9 & 6.9 \\\hline
Trial 6 & 2.3 & 6.3 \\\hline
Trial 7 & 2.3 & 9.7 \\\hline
Overall & 2.3 & 11.1 \\\hline
\end{tabular}}
\end{table}

The experimental results showed that path following accuracy was relatively consistent in all trials and showed close values of positional errors between trot and gallop gaits. However, in this gait, the systematic error can again be observed, suggesting that the UAV's inclination from the trajectory may occur as a result of the initial angle calibration of the limbs.
\subsubsection{Discussion}
The experimental results revealed sufficient accuracy of the linear path following in several gaits, with trot and gallop gaits being more beneficial for CoM of MorphoGear straight line following. Moreover, there was a deviation of the robot's trajectory to the side, which indicates the requirement for a precise initial calibration of the robot. This problem may also occur at the mechanical level and require additional backlash compensation. It is hypothesized that this error occurred due to a backlash in the fasteners of the parts as well as an error in the calibration of the initial angle $\theta_0$.

\begin{figure}[h!]
\centering
\includegraphics[width=0.9\linewidth]{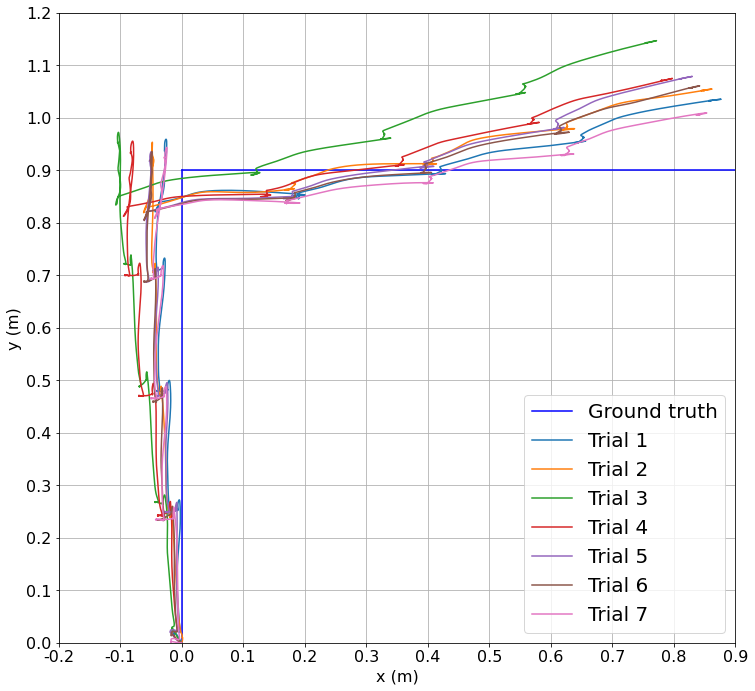}
\caption{Right angle trajectory experiment with trot gait motion of the MorphoGear.}
\label{fig_type4}
\end{figure}

Additionally, a test with a right angle trajectory was carried out. In this test, the robot took four steps with the trot gait, then took four steps to the right with the same gait without rotation of the body. Trajectories of the robot's motion with the trot gait are shown in Fig.~\ref{fig_type4}.

Table \ref{Table Result 4} shows the RMSE of 7.0 cm and the maximal error of 20.7 cm after angular path following. In this test, low deviations to the left on both the forward and the right sections of the trajectory were observed, indicating the initial calibration problem being a more probable issue than the mechatronic design or a backlash in the joints of the robot.

\begin{table}[h!]
\centering{
\caption{Path Following Error \\ for Angular Line Trajectory and Trot Gait}
\setlength{\tabcolsep}{6pt} 
\renewcommand{\arraystretch}{1}
\begin{tabular}{ | c | c | c | }
\hline
\label{Table Result 4}
\begin{tabular}{p{1.5cm}cp{4cm}} \end{tabular} & {RMSE, cm} & {Max error, cm} \\\hline
Trial 1 & 5 & 11.3 \\\hline
Trial 2 & 6.9 & 14.6 \\\hline
Trial 3 & 10.2 & 19.6 \\\hline
Trial 4 & 8 & 16.8 \\\hline
Trial 5 & 6.8 & 20.7 \\\hline
Trial 6 & 6.7 & 14.8 \\\hline
Trial 7 & 5.4 & 11.9 \\\hline
Overall & 7 & 20.7 \\\hline
\end{tabular}}
\end{table}

\section{Conclusion}
We have developed a novel MorphoGear UAV with morphogenetic gear for terrestrial locomotion, object grasping, and aerial motion. The selected parameters of the walking behavior allowed us to achieve stable robot motion, with feedback from the servos and tracking data from the Vicon mocap system. The experimental analysis of three walking gaits revealed the lowest positional error (RMSE of 1.9 cm and maximal error of 8.2 cm) with a trot gait. The second type of gait causes the platform to sway (maximal error of 21.1 cm), but can be adapted for uneven surfaces and potentially stabilized with the help of air engines in future work. In addition, the experiments revealed a low systematic deviation of the robot from the trajectory to the side, indicating the need to calibrate the robot and its initial positions.
In the future, we are planning to explore the navigation of MorphoGear taking into account its dynamics. Additionally, manipulation scenarios with MorphoGear on the ground and in midair will be explored. Our goal will be to evaluate the ability of the robot to learn through DL methods which limbs to use for grasping and which limbs (and rotors) to use for stabilization in complex manipulation tasks.


%




\ifCLASSOPTIONcaptionsoff
 \newpage
\fi



%


\bibliographystyle{IEEEtran}
\bibliography{bare_jrnl}

%








\end{document}